\documentclass{article}
\usepackage{spconf,amsmath,graphicx}
\usepackage{amssymb} 
\usepackage{caption}
\usepackage{subcaption}

\title{Subspace Representation Learning for Few-shot Image Classification}
\name{Ting-Yao Hu, Zhi-Qi Cheng, and Alexander G. Hauptmann }
\address{Carnegie Mellon University, Pittsburgh, PA, USA}
\begin{document}
%
\maketitle
\begin{abstract}
In this paper, we propose a subspace representation learning (SRL) framework to tackle few-shot image classification tasks.
It exploits a subspace in local CNN feature space to represent an image, and measures the similarity between two images according to a weighted subspace distance (WSD).
When $K$ images are available for each class, we develop two types of template subspaces to aggregate $K$-shot information: the prototypical subspace (PS) and the discriminative subspace (DS).
Based on the SRL framework, we extend metric learning based techniques from vector to subspace representation.
While most previous works adopted global vector representation, using subspace representation can effectively preserve the spatial structure, and diversity within an image.
We demonstrate the effectiveness of the SRL framework on three public benchmark datasets: MiniImageNet, TieredImageNet and Caltech-UCSD Birds-200-2011 (CUB), and the experimental results illustrate competitive/superior performance of our method compared to the previous state-of-the-art.

\end{abstract}
%
%

\section{Introduction}
\label{sec:intro}
Deep neural networks (DNN) has enabled huge advances in many computer vision tasks, such as image classification \cite{russakovsky2015imagenet} and object detection \cite{fasterrcnn}. 
However, the astounding success of DNN is conditioned on the availability of large scale datasets with thorough manual annotation, which is usually too expensive for real-world applications.
In contrast, the human visual system is capable of learning a new visual concept with only a few annotated examples.
This phenomenon inspired the development of few-shot learning \cite{snell2017prototypical,finn2017model,lee2019meta,zhang2020deepemd,li2019revisiting,Sachin2017,vinyals2016matching,relationnet,ye2020few}, which aims at obtaining a reliable prediction model that can easily be generalized to unseen concepts.

One of the mainstream approaches to few-shot image classification is based on metric learning \cite{snell2017prototypical,vinyals2016matching,relationnet,ye2020few}.
This type of method focuses on learning a good metric function from known concepts with sufficient labeled data, and transferring the learned metric to unseen concepts.
Specifically, they exploit a Convolutional Neural Network (CNN) to extract a feature vector for each image, and measure the similarity between two images in hidden feature space based on distance functions, such as Euclidean and cosine distance.
While receiving state-of-the-art performance, metric learning based methods are still not able to handle some unseen visual concepts with large intra-class variation and cluttered background \cite{zhang2020deepemd}.
One major reason is that an image-level feature vector ignores the spatial structure and diversity of an image.
In the context of few-shot image classification, this problem becomes more severe since there is not enough supervision signal to guide the network focusing on the correct local regions.

In this paper, we propose a novel subspace representation learning (SRL) framework for few-shot image classification tasks.
The SRL framework represents an image as a subspace extracted from its local CNN features.
Then, the similarity between two images is measured by a subspace-to-subspace distance.
While many strategies \cite{zuccon2009semantic} have been proposed to compute the distance between two subspaces, we choose the weighted subspace distance (WSD) \cite{li2009weighted}, which considers the importance of each dimension and reflects the original distribution of local CNN features. 
The SRL framework supports end-to-end training using a loss function related to distance based classifier \cite{chen2019closer}.
When $K$-shot examples are available for a class, SRL utilizes a template subspace to summarize the information from $K$ images.
To do so, we adapt two popular strategies developed for vector space to SRL framework.
The first one is to obtain a class-specific subspace prototype by calculating the average of subspaces, while the second one is to learn a set of task-specific discriminative subspaces.
Both strategies can be formulated as an optimization problem on a Stiefel manifold.
In comparison with an image-level vector, a subspace is a compact representation capable of capturing the spatial structure and diversity of an image.



To evaluate the proposed SRL framework, we conduct experiments on three popular benchmarks for few shot image-classification: MiniImageNet, TieredImageNet and Caltech-UCSD Birds-200-2011 (CUB).
Experimental results show that our method achieves competitive/superior performance compared to state-of-the-art few-shot learning approaches.
In summary, our main contributions are three fold:
\begin{itemize}
    \item we tackle the few-shot image classification task by proposing the idea of subspace representation. \\
    \vspace{-0.1in}
    \item we propose and compare two types of template subspace to aggregate $K$-shot information. \\
    \vspace{-0.1in}
    \item our SRL framework achieves state-of-the-art performance on three public benchmarks. \\
    \vspace{-0.1in}
\end{itemize}

\section{Related Work}
\label{sec:related_work}
Previous approaches to few-shot image classification can be roughly divided into three categories:
(1) Distance metric based methods \cite{snell2017prototypical,vinyals2016matching,relationnet} construct a proper hidden feature space, whose distance metric is used to determine the image-class or image-image similarity.
The distance metric can be a non-parametric function \cite{snell2017prototypical,vinyals2016matching} or a parametric network module \cite{relationnet}.
(2) Optimization based methods \cite{finn2017model,jamal2019task,munkhdalai2018rapid} aim at learning a good initialization for the model so that it can be quickly fine-tuned to a target task with limited amount of data.
(3) Hallucination based methods \cite{hariharan2017low,wang2018low,lin2019semantics,kimmodel} solve data scarcity by generating more training samples.
The generation process is done in either hidden feature space or raw image space.
While achieving promising performance, these methods adopt a global feature vector to represent an image.

As an extension of distance metric based methods, some recent works \cite{zhang2020deepemd,li2019revisiting} rely on a local feature set representation to preserve the spatial structure of an image, and define their own metric to measure the similarity between two feature sets.
Li et al. \cite{li2019revisiting} calculates cosine distance between local feature pairs, and aggregates them by a k-NN classifier.
Zhang et al. \cite{zhang2020deepemd} adopt earth mover's distance (EMD) to discover an optimal matching between local feature sets.
In comparison, our method extracts a subspace representation for each local CNN feature set, and computes the weighted subspace distance between two sets.

The concept of subspace learning has been utilized to solve few-shot image classification \cite{pmlr-v97-yoon19a,simon2020adaptive}.
However, these works consider the subspace in a global, image-level feature space.
Also, they make use of a subspace projection operation to compute subspace-point distance and measure class-to-image similarity.
In comparison, our method represents an image as a subspace in local feature space, and calculates subspace-subspace distance.

\section{Proposed Method}
\label{sec:method}

The goal of few-shot image classification task is to build a prediction model that can be quickly adapted to unseen classes with limited amount of annotated examples.
Most previous works validate their approaches to this task using $N$-way $K$-shot classification as testing scenario.
Specifically, we are given a support set $S= \{\{(x_{i, j}, y_i) \}_{j=1}^{K} \}_{i=1}^N$ and query set $Q = \{(x_q, y_q)\}_{q=1}^{N_q}, y_q \in \{y_i\}_{i=1}^N$, where $(x,y)$ is the pair of raw image and class label.
The prediction model is trained/adapted based on $S$, and evaluated on the classification results of $Q$.

\begin{figure*}
\centering
  \includegraphics[width=0.9\textwidth]{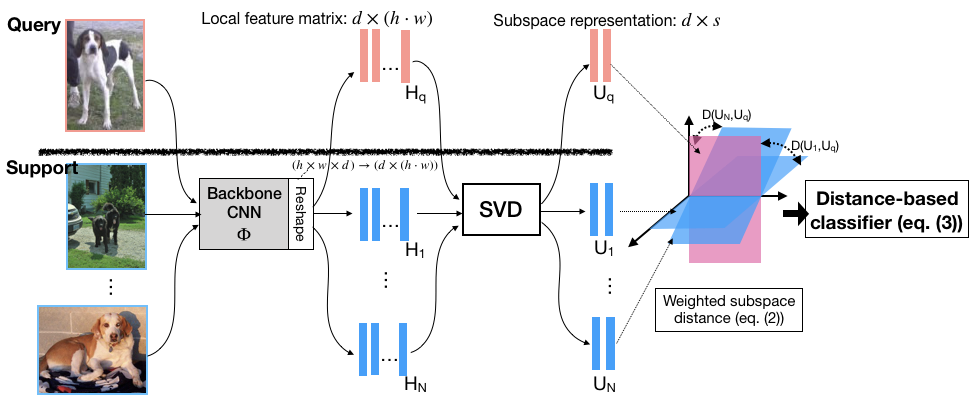}
  \caption{The overall architecture of proposed subspace representation learning (SRL) framework. The backbone CNN extracts the local feature map ($h \times w \times d$) from each query and support image. After reshaping the feature map into a matrix $H \in \mathbb{R}^{d \times (h \cdot w)}$, whose columns are the CNN features at every spatial location, we extract the subspace representation by conducting SVD. The similarity between two images is determined by a weighted subspace distance (WSD). The end-to-end training is guided by the loss function of a distance based classifier.}
  \label{fig:arch}
\end{figure*}

In this work, we propose subspace representation learning (SRL) framework to tackle this task.
The overall architecture is illustrated in Fig. \ref{fig:arch}.
Concretely, our method represents an image as a subspace, which is estimated from the reconstruction of local CNN features of this image.
The dis-similarity between two images can be determined by a weighted subspace distance (WSD) between two subspaces.
In the rest of this section, we first introduce the concept of subspace representation.
Then, we elaborate the WSD adopted in SRL framework, and the end-to-end training process in the context of few-shot image classification.
Finally, we describe two types of template subspace, which can effectively summarize information about a specific class from K-shot examples.

\subsection{Subspace Representation}
Our method exploits a subspace to represent each training/testing image.
Given an image $x$, we extract the hidden feature map ($h \times w \times d$ tensor) using a backbone CNN with parameter $\Phi$, and collect the $d$-dimensional local feature vectors at all the spatial locations to form a matrix $H \in \mathbb{R}^{d \times (h \cdot w)}$.
Then, our method finds the best-fit $s$-dimensional subspace $U \in \mathbb{R}^{d \times s}$, which minimizes the reconstruction error with respect to $H$:
\begin{equation}
\begin{split}
    \min_{U}\; &||H - UU^TH||_F \\
    s.t. \;& U^TU = I
\end{split}
\label{eq:sub_rep_problem}
\end{equation}
where $||\cdot||_F$ is the Forbeneous norm of a matrix.
This optimization problem can be solved by singular value decomposition (SVD) of $H$, and the optimal $U$ is obtained by the top-$s$ right-singular vectors with the largest singular values.

Similar to other works \cite{li2019revisiting,zhang2020deepemd} leveraging local CNN features, our SRL framework is able to capture the spatial structure of an image.
However, there are two additional reasons why we choose to construct a  subspace representation.
First, a subspace encourages the preservation of diversity because of the orthonormal constraint in eq. (\ref{eq:sub_rep_problem}).
Second, using a subspace results in a compact representation for image, since we can set a small $s$ without sacrificing the performance.
More analysis of these two properties will be elaborated in the experiment section.

\subsection{Weighted Subspace Distance}
To conduct metric learning in the space of subspace representation, we need to calculate subspace-to-subspace distance.
While several types of distance have been proposed, our SRL framework utilizes the weighted subspace distance (WSD) introduced in \cite{li2009weighted}:
Given two subspaces $U_1$ and $U_2$ representing two images $x_1$ and $x_2$, WSD is expressed as:
\begin{equation}
\begin{split}
    D(U_1, U_2) & = \sqrt{1 - \sum_{i=1}^s \sum_{j=1}^s \sqrt{ \lambda_{1,i}' \lambda_{2,j}' } (u_{1,i}^Tu_{2,j})^2 } \\
    \lambda_{1,i}' & = \frac{ \lambda_{1,i} }{ \sum_{l=1}^{s} \lambda_{1,l}}, \; \lambda_{2,j}'  = \frac{ \lambda_{2,j} }{ \sum_{l=1}^{s} \lambda_{2,l}}
\end{split}
\label{eq:wsd}
\end{equation}
where $u_{1,i(2,j)}$ and $\lambda_{1,i(2,j)}$ are the $i(j)$-th column of $U_{1(2)}$, and the corresponding singular value obtained from SVD operation, respectively.
Comparing to other types of subspace distance, WSD considers the relative importance of each basis component in a subspace, so it can better capture the distribution of original local feature set.

\subsection{End-to-End Training}
The training process of our SRL framework follows the episodic learning mechanism \cite{vinyals2016matching}, which mimics the situation of a testing phase.
In each training iteration, we sample a pair of support and query sets $(S,Q)$ from training dataset.
Then, we extract the subspace representation for every image in $S$ and $Q$, and plug the WSD (eq. (\ref{eq:wsd})) between the two subspaces into a distance based classifier \cite{chen2019closer}.
Thus, the training objective of SRL can be formulated as: 
\begin{equation}
    \mathcal{L}_{e2e}(\Phi) = \sum_q log \left( \frac{exp(-D(U_{y_q}, U_q))}{\sum_{i=1}^N exp(-D(U_{y_i},U_q))} \right)
\label{eq:obj}
\end{equation}
where $U_q$ is the subspace representation of query. $U_{y_i}$ is the subspace representation of the supported image of class $y_i$ if $K=1$. 
In the case of $K>1$, $U_{y_i}$ stands for the template subspace that summarizes the $K$-shot information from class $y_i$.
The estimation of template subspace will be discussed in Sec. \ref{sec:kshot}.
By minimizing eq. (\ref{eq:obj}), the parameter set $\Phi$ of backbone CNN can be learned in an end-to-end manner.

\subsection{Template Subspace for K-shot Learning}
\label{sec:kshot}

\begin{figure*}[t]
\centering
  \begin{subfigure}[b]{0.35\textwidth}
         \centering
         \includegraphics[width=\textwidth]{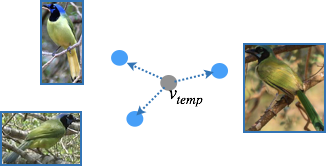}
         \caption{Prototypical network \cite{snell2017prototypical}}
         \label{fig:kshot_proto}
     \end{subfigure}
     \hspace{5em}
     \begin{subfigure}[b]{0.35\textwidth}
         \centering
         \includegraphics[width=\textwidth]{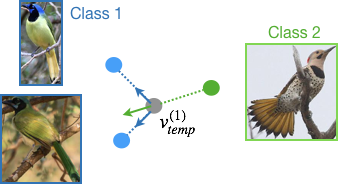}
         \caption{Distance-based classifier. \cite{chen2019closer}}
         \label{fig:kshot_dist}
     \end{subfigure}
     \par\bigskip
     \begin{subfigure}[b]{0.35\textwidth}
         \centering
         \includegraphics[width=\textwidth]{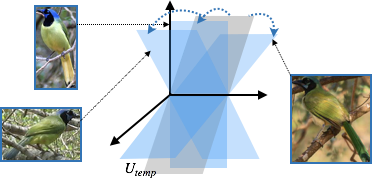}
         \caption{Prototypical subspace (PS). (eq. (\ref{eq:subspace_proto}))}
         \label{fig:kshot_proto_subspace}
     \end{subfigure}
     \hspace{5em}
     \begin{subfigure}[b]{0.35\textwidth}
         \centering
         \includegraphics[width=\textwidth]{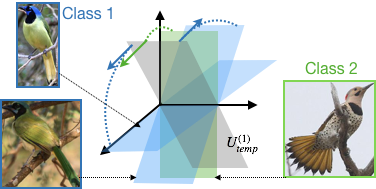}
         \caption{Discriminative subspace (DS). (eq. (\ref{eq:subspace_disc}))}
         \label{fig:kshot_disc_subspace}
     \end{subfigure}
     \caption{Comparison among different strategies for template vector/subspace ($v_{temp}$/$U_{temp}$) extraction from $K$-shot information. (a) A prototypical network takes mean vector of $K$-shot vector representations. (b) A distance based classifier obtains the class template vector whose Euclidean/Cosine distances to $K$-shot examples minimize a cross-entropy loss. (c) A prototypical subspace (PS) is the average subspace of $K$-shot subspaces. (d) A discriminative subspace (DS) optimizes a distance based classifier with WSD.}
     \label{fig:kshot}
\end{figure*}

To cope with the $K$-shot learning scenario, the SRL framework computes a template subspace $U_{temp}$ to aggregate the information from $K$ subspaces for each class.
In this work, we propose two types of template subspace to represent a class: A prototypical subspace and a discriminative subspace.

The prototypical subspace (PS) is the "average" of all $K$ subspaces, following the spirit of ProtoNet \cite{snell2017prototypical} in vector space.
Specifically, given K subspaces $U_1,U_2,...,U_K$ representing $K$ images of the same class, the prototypical subspace is obtained by the minimizing the summation of the distances between $U_{temp}$ and $U_i$:
\begin{equation}
    \mathcal{L}_{ps}(U_{temp}) = \sum_{j=1}^K D(U_{temp},U_j)
\label{eq:subspace_proto}
\end{equation}
On the other hand, the discriminative subspace (DS) can be calculated by training a distance-based classifier with respect to a support set $S$ of a $N$-way, $K$-shot classification task.
Given the set of $NK$ subspaces, $\{\{U_{i, j} \}_{j=1}^{K} \}_{i=1}^N$, extracted from $S$, the set of $N$ template subspaces $\boldsymbol{U_{temp}}= \{U_{temp}^{(i)}\}_{i=1}^N$ for N classes would be the minimizer of the following loss function:
\begin{equation}
    \mathcal{L}_{ds}(\boldsymbol{U_{temp}}) = \sum_{i=1}^N \sum_{j=1}^K log \left( \frac{exp(-D(U_{i,j}, U_{temp}^{(i)}))}{\sum_{l=1}^N exp(-D(U_{i,j}, U_{temp}^{(l)}))} \right)
\label{eq:subspace_disc}
\end{equation}
Please note that $\{U_{temp}^{(i)}\}_{i=1}^N$ are task-specific, since they are derived jointly from a $N$-way, $K$-shot task.
In contrast, PS is class-specific because the optimal $U_{temp}$ in eq. (\ref{eq:subspace_proto}) is independent to other $(N-1)K$ images in $S$.
Fig. \ref{fig:kshot} compares PS and DS, along with their correspondence in vector space.

While minimizing $\mathcal{L}_{ps}$ and $\mathcal{L}_{ds}$, $U_{temp}$ is subject to the orthonormal constraint ($U^TU=I$), which prevents a closed-form solution of both problems.
To solve this type of optimization problem with a SGD-like algorithm, we exploit the Cayley transform \cite{nishimori2005learning}, projecting the gradient to the tangent space of a Stiefel manifold.
We describe the update rule of estimating PS as an example.
Let $Z=\partial \mathcal{L}_{ps}/ \partial U_t$, where $U_t$ is the current estimated $U_{temp}$ in eq. (\ref{eq:subspace_proto}).
The calculation of the next $U_{temp}$ estimation $U_{t+1}$ can be expressed as:
\begin{equation}
\begin{split}
    &W = \hat{W} - \hat{W}^T, \; \hat{W} = ZU_t - \frac{1}{2}U_t(U_t^TZU_t^T)\\
    &U_{t+1} = (I-\frac{\alpha}{2}W)^{-1}(I+\frac{\alpha}{2}W)U_{t}
\end{split}
\label{eq:cayley}
\end{equation}
where $\alpha$ is a hyper-parameter analogous to the learning rate in SGD-like algorithms.


\section{Experiment}
\label{sec:ep}
\begin{table*}[t]
\centering
\begin{tabular}{l|ll|ll}
\hline
                            &  \multicolumn{2}{l|}{MiniImageNet}       & \multicolumn{2}{l}{TieredImageNet}      \\ \hline
Methods                      & 5-way, 1-shot      & 5-way, 5-shot      & 5-way, 1-shot      & 5-way, 5-shot      \\ \hline
Baseline++ \cite{chen2019closer}           & 53.97 $\pm$ 0.79\% & 75.90 $\pm$ 0.61\% & 61.49 $\pm$ 0.51\% & 82.37 $\pm$ 0.67\% \\
ProtoNet$^*$ \cite{snell2017prototypical}      & 63.56 $\pm$ 0.34\% & 81.08 $\pm$ 0.18\% & 69.53 $\pm$ 0.36\% & 84.02 $\pm$ 0.23\% \\
MetaOpt-SVM \cite{lee2019meta}                       & 62.64 $\pm$ 0.82\% & 78.63 $\pm$ 0.46\% & 65.99 $\pm$ 0.72\% & 81.56 $\pm$ 0.53\% \\
MatchNet \cite{vinyals2016matching}           & 63.08 $\pm$ 0.80\% & 75.99 $\pm$ 0.60\% & 68.50 $\pm$ 0.92\% & 80.60 $\pm$ 0.71\% \\
DSN-MR \cite{simon2020adaptive}                        & 64.60 $\pm$ 0.62\% & 79.51 $\pm$ 0.50\% & 67.39 $\pm$ 0.82\% & 82.85 $\pm$ 0.56\% \\
FEAT \cite{ye2020few}              & 66.78 $\pm$ 0.20\% & 82.05 $\pm$ 0.15\% & 70.80 $\pm$ 0.23\% & 84.79 $\pm$ 0.16\% \\
Seq-distill  \cite{seqdistill}                     & 64.80 $\pm$ 0.60\% & 82.14 $\pm$ 0.43\% & 71.52 $\pm$ 0.69\% & 86.03 $\pm$ 0.49\% \\
DN4$^{* \dagger}$  \cite{li2019revisiting}           & 63.72 $\pm$ 0.32\% & 81.54 $\pm$ 0.20\% & 70.23 $\pm$ 0.33\% & 84.01 $\pm$ 0.24\% \\
DeepEMD$^{\dagger}$ \cite{zhang2020deepemd}                           & 65.91 $\pm$ 0.82\% & 82.43 $\pm$ 0.56\% & 71.16 $\pm$ 0.87\% & 86.03 $\pm$ 0.58\% \\
SRL, DS (ours)$^\dagger$                    & \textbf{67.00 $\pm$ 0.27\%} & \textbf{82.68 $\pm$ 0.18\%} & \textbf{71.88 $\pm$ 0.32\%} & \textbf{86.24 $\pm$ 0.22\%} \\ \hline
\end{tabular}
\caption{Results on MiniImageNet and TieredImageNet. All the methods use ResNet-12 as the backbone network. For SRL, we set subspace basis size, $s=5$. *: Our re-implementation. $\dagger$: Using local CNN feature.}
\label{tab:sota}
\end{table*}

\subsection{Implementation}
For a fair comparison, we select the commonly used ResNet-12 as the backbone network of our SRL framework.
The softmax layer and the spatial average pooling are removed from the backbone.
All the images are resized to $84 \times 84$ pixels, and become $5 \times 5 \times 640$ tensors after analyzed by the backbone network.
To optimize the parameters of this backbone ResNet-12, we conduct a two-step training process.
In the first step, we pre-train the parameters by minimizing a cross entropy loss function of a standard classification task using training classes.
In the second step, we perform the episodic training mechanism described in Sec. \ref{sec:method}.
We adopt SGD optimizer for 10k iterations with an initial learning rate $0.002$, which decreases by a factor $0.1$ for every 2k iteration.
No data augmentation methods are applied during the episodic training.

The PS and DS are initialized by the subspaces extracted from the union of $K$ local CNN feature sets.
Then, we update these template subspaces using SGD with Cayley transform for 50 iterations.
The learning rate $\alpha$ for PS and DS are $0.1$ and $0.01$, respectively.

In our experiments, we follow the standard 5-way, 1-shot and 5-shot classification protocols, and sample 5,000 tasks with 15 query images for each target class.
The category of each query image is predicted independently (inductive scenario).
We report the average accuracy with the $95\%$ confidence interval of all the sampled tasks.

\subsection{Dataset}
To evaluate the efficacy of SRL, we conduct experiments on three benchmark datasets:
MiniImageNet \cite{Sachin2017}, TieredImageNet \cite{ren18fewshotssl} and Caltech-UCSD Birds-200-2011 \cite{cub}.

MiniImageNet is a subset of ImageNet \cite{russakovsky2015imagenet}. 
It consists of 100 classes of images, and 600 images per class.
The 100 classes are divided into 64, 16, 20 for training, validation and testing sets, respectively.

TieredImageNet contains 608 classes from 34 super-class of ImageNet, and 779,165 images in total.
The set of 608 classes is divided into subsets with 351, 97, and 160 classes for model training, validation, and testing, respectively, according to their super-class.
This arrangement increases the domain gap between training and evaluation phase.

Caltech-UCSD Birds-200-2011 (CUB) was designed for fine grained image recognition.
It contains 200 classes of bird images, 11,788 images in total.
Following the setup in previous works \cite{ye2020few,zhang2020deepemd}, we split the 200 classes into 100, 50, 50, classes for model training, validation and testing, respectively.
Comparing to other two aforementioned datasets, CUB is challenging because of the subtle difference among bird types. 

\begin{table}[t]
\centering
\begin{tabular}{l|ll}
\hline
Methods                      & 5-way, 1-shot               & 5-way, 5-shot               \\ \hline
ProtoNet$^*$      & 72.45$\pm$0.34\%          & 85.94$\pm$0.23\%          \\
MatchNet                      & 71.87$\pm$0.85\%          & 85.08$\pm$0.57\%          \\
Baseline++                      & 69.55$\pm$0.89\%          & 85.17$\pm$0.50\%          \\
DN4$^{* \dagger}$            & 72.30$\pm$0.32\%          & 85.23$\pm$0.23\%          \\
DeepEMD$^{\dagger}$                           & \textbf{75.65$\pm$0.83\%} & 88.69$\pm$0.50\%          \\
SRL, DS (ours)$^{\dagger}$                        & 75.32$\pm$0.27\%         & \textbf{88.81$\pm$0.21\%} \\ \hline
\end{tabular}
\caption{Results on CUB. All the methods use ResNet-12 as the backbone network. For SRL, we set subspace basis size, $s=5$. *: Our re-implementation. $^\dagger$: Using local CNN feature.}
\label{tab:sota_cub}
\end{table}

\begin{figure}[t]
\centering
  \begin{subfigure}[b]{0.4\textwidth}
         \centering
         \includegraphics[width=\textwidth]{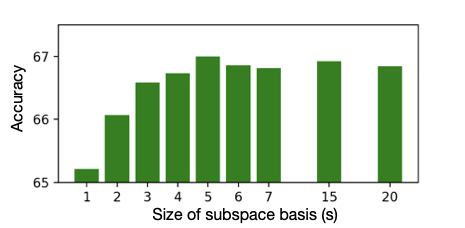}
         \caption{Results on 5-way, 1-shot task of MiniImageNet}
         \label{fig:kshot_proto}
     \end{subfigure}
     \begin{subfigure}[b]{0.4\textwidth}
         \centering
         \includegraphics[width=\textwidth]{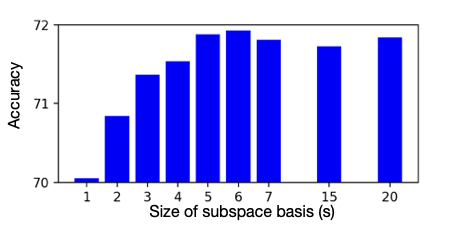}
         \caption{Results on 5-way, 1-shot task of TieredImageNet}
         \label{fig:kshot_dist}
     \end{subfigure}
     \caption{Sensitivity analysis with respect to the size of subspace basis. The results show that the accuracy saturates around $s=6$.}
     \label{fig:basis_size}
\end{figure}

\subsection{Analysis for SRL Design}
To better understand the property of SRL and validate our design choices, we conduct two quantitative studies: (1) sensitivity analysis for basis size of subspace representation (2) validating the choice of WSD.

In the first study, we adjust subspace basis size $s$, which is also the number of columns of $U$ in eq. (\ref{eq:sub_rep_problem}), and report the performance of 5-way, 1-shot classification task in MiniImageNet and TieredImageNet.
From the results shown in Fig \ref{fig:basis_size}, we find that the performance gets better when $s$ becomes larger, but saturates around $s=6$.
This observation is expected since the later included subspace basis components are with smaller singular values, and less important according to the definition of WSD.
The results also suggest that subspace representation with basis size $s=6$ is enough to preserve essential information for few-shot classification task.

\begin{figure}[t]
    \centering
    \includegraphics[width=0.85\linewidth]{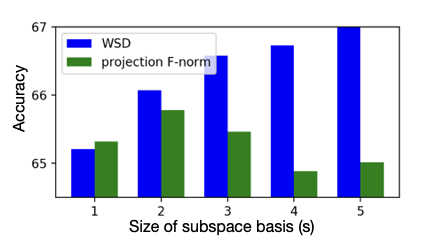}
    \caption{Comparison between SRL implementations with two types of subspace distance: WSD (eq. (\ref{eq:wsd})) and projection F-norm (eq. (\ref{eq:pfnorm}))}
    \label{fig:sd}
\end{figure}

In the second study, we compare the adopted WSD with another commonly used subspace distance measure, projection F-norm \cite{simon2020adaptive,edelman1998geometry}:
\begin{equation}
    D_p(U_1, U_2) = ||U_1U_1^T-U_2U_2^T||_F^2 = 2s-2||U_1^TU_2||_F^2
\label{eq:pfnorm}
\end{equation}
Specifically, we replace WSD with $D_p(U_1, U_2)$ in SRL, and follow the same training procedure to optimize the backbone CNN.
The performance of SRL with these two subspace distance functions are compared on the 5-way, 1-shot task of MiniImageNet.
From the results illustrated in Fig. \ref{fig:sd}, we observe that the performance of SRL with projection F-norm (eq. (\ref{eq:pfnorm})) drops while $s$ increasing, and performs worse than SRL with WSD in general.
The potential reason is that WSD re-weights the importance of each basis component, reflecting the distribution of local CNN features.

\subsection{Analysis for Template Subspace}
\begin{table}[t]
\begin{tabular}{l|ll}
\hline
                 & MiniImageNet       & TieredImageNet     \\ \hline
SRL, PS               & 82.14 $\pm$ 0.18\% & 85.86 $\pm$ 0.23\% \\
SRL, DS               & \textbf{82.68 $\pm$ 0.18\%} & \textbf{86.24 $\pm$ 0.22\%} \\
Baseline (Union) & 80.93 $\pm$ 0.20\% & 83.98 $\pm$ 0.23\% \\
Baseline (NN)    & 80.56 $\pm$ 0.23\% & 83.11 $\pm$ 0.26\% \\ \hline
\end{tabular}
\caption{Comparison among K-shot aggregation methods on 5-way, 5-shot task of MiniImageNet and TieredImageNet.}
\label{tab:kshot}
\end{table}

In this experiment, we compare two types of template subspace, PS and DS, along with two other naive methods, baseline (union) and baseline (NN), in K-shot learning scenario.
Baseline (union) extracts the subspace from the union of all the local CNN features from K-shot images.
It is also adopted as the initialization step of PS and DS.
Baseline (NN) computes the subspace-subspace distance from query image to the nearest neighbor support image in terms of WSD.
All the methods are evaluated on the 5-way, 5-shot classification task of MiniImageNet and TieredImageNet.

From the results illustrated in Table \ref{tab:kshot}, we can see that both PS and DS can outperform two naive baselines, and DS receives the best performance.
A possible explanation is that DS is optimized based on task-specific information, while PS is only conditioned on the intra-class information. 
Thus, when two confusing unseen concepts appear in the same sampled task, DS has a better chance to distinguish them.

\subsection{Comparison with State-of-the-art}
We compare the performance of our SRL framework with two approaches using local feature set, DeepEMD \cite{zhang2020deepemd} and DN4 \cite{li2019revisiting}, as well as other previous state-of-the-art methods.
The experiment results are summarized in Table \ref{tab:sota} and \ref{tab:sota_cub}.
From this set of results, we have the following observations.
First, ProtoNet from our implementation receives competitive performance on three datasets, serving as a strong baseline.
Second, for 1-shot, 5-way task, our proposed SRL performs the best on MiniImageNet and TieredImageNet, and is only slightly worse than DeepEMD on CUB dataset.
Third, SRL outperforms all the other methods on the 5-way, 5-shot task of all three datasets.

\subsection{Visualization}
\begin{figure}[t]
    \centering
    \includegraphics[width=0.85\linewidth]{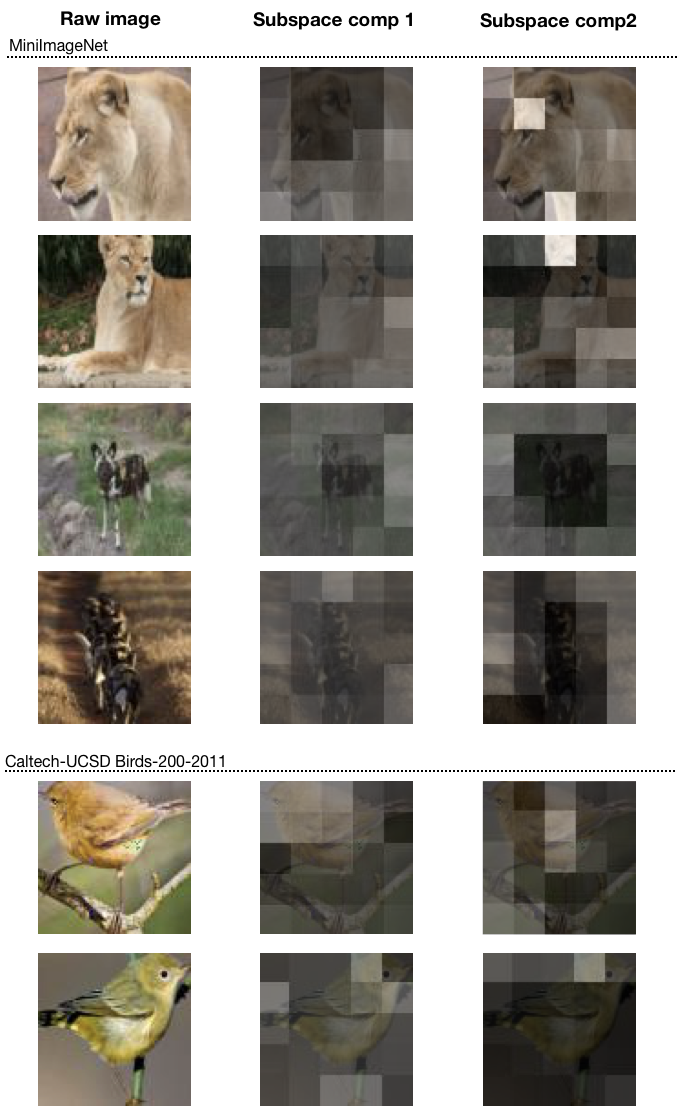}
    \caption{Visualization of subspace representation. Raw images are from MiniImageNet and CUB dataset. Brighter regions indicate higher cosine similarity between subspace basis component and the local CNN feature.}
    \label{fig:vis}
\end{figure}

To understand the underline information captured by subspace representation, we visualize the subspace basis components extracted from images of MiniImageNet and CUB datasets.
Specifically, we calculate the cosine similarity between each basis component and the local CNN feature of each $5 \times 5$ regions.
Fig. \ref{fig:vis} shows the visualization results of the first two basis components. 
The brightness of each spatial region is proportional to the cosine similarity between the components and the local CNN feature.
From the results, we can see that the first component contains shared information among all the features, while the second focuses on some specific local regions.
These local regions reflect the characteristics of the corresponding class, which would be useful for classification task.
However, observing the third and fourth example images from MiniImageNet, we find that their second basis components are highly correlated to the background regions
It indicates that the proposed SRL sometimes suffers from some dataset bias, and fails to represent the object.

\section{Conclusion}
\label{sec:conclusion}
In this paper, we propose a subspace representation learning (SRL) framework for few shot image classification task.
It represents an image as a subspace in local CNN feature space, and compares two images by calculating a weighted subspace distance (WSD).
When $K$-shot information is available, we propose two types of class template representation for SRL: a prototypical subspace (PS) and a discriminative subspace (DS).
The estimation of PS and DS can be formulated as an optimization problem in a Stiefel manifold.
The experiment results on three public benchmark datasets show that our SRL framework achieves state-of-the-art performance.

The proposed SRL framework extends the concept of deep metric learning from vector to subspace, and serves as a general tool for modeling the local structure of an image
It can easily be applied to other end-to-end learning network architectures.
Our future work will focus on exploring the applicability of this approach to other tasks and domains.



\bibliographystyle{IEEE}
\bibliography{refs}

\end{document}